\pdfoutput=1

\documentclass[11pt]{article}

\usepackage[final]{acl}

\usepackage{times}
\usepackage{latexsym}

\usepackage[T1]{fontenc}

\usepackage[utf8]{inputenc}

\usepackage{microtype}

\usepackage{inconsolata}

\usepackage{graphicx}

\title{Improved Methods for Model Pruning and Distillation}

\author{Wei Jiang \\
  Suanfamama \\
  \texttt{wei@suanfamama.com} \\\And
  Anying Fu \\
  Suanfamama \\
  \texttt{anying@suanfamama.com} \\\And
  Youling Zhang \\
  Suanfamama \\
  \texttt{youling@suanfamama.com} \\}

\begin{document}
\maketitle
\begin{abstract}
Model pruning is a performance optimization technique for large language models like R1 or o3-mini. However, existing pruning methods often lead to significant performance degradation or require extensive retraining and fine-tuning. This technique aims to identify and remove neurons, connections unlikely leading to the contribution during the human-computer interaction phase. Our goal is to obtain a much smaller and faster knowledge distilled model that can quickly generate content almost as good as those of the unpruned ones. We propose MAMA Pruning, short for Movement and Magnitude Analysis, an improved pruning method that effectively reduces model size and computational complexity while maintaining performance comparable to the original unpruned model even at extreme pruned levels. The improved method is based on weights, bias fixed in the pre-training phase and GRPO rewards verified during the post-training phase as our novel pruning indicators. Preliminary experimental results show that our method outperforms and be comparable to state-of-the-art methods across various pruning levels and different downstream computational linguistics tasks.
\end{abstract}

\section{Introduction}

Large language models face significant computational challenges due to massive model sizes and the high query loads that these systems need to support. These models, along with related large-scale production systems, are responsible for processing and integrating vast amounts of data, including web pages, videos, and multimodal content into underlying network architectures such as Transformers and Diffusion models \cite{fei2025vision}.

One crucial cost factor is the query processing per user, which must scale with both data size and query load. As a result, large foundational models devote substantial hardware and energy resources to this kind of generation task. There has been extensive research on improving query processing performance, including work on various caching techniques, retrieval information systems, and high-performance knowledge representation. To address these challenges, a significant number of optimization techniques, commonly referred to as model pruning and distillation, have emerged to enhance the efficiency and effectiveness of the generation processes \cite{jeff2025vision}.

In this paper, we propose an improved model pruning algorithm based on novel indicators derived from an in-depth analysis of weights, biases, activations and rewards. Our method significantly enhances the performance and efficiency of large language models. We also want to demonstrate through extensive experiments that our approach outperforms existing state-of-the-art pruning techniques across various evaluation metrics \cite{wei2024vision}.

\section{Proposed Methods for pruning}

MAMA Pruning is grounded in a systematic approach that identifies and preserves dynamically significant weights by redistributing less important weights to more critical connections within the network. This ensures the maintenance of overall information flow and network adaptability, even under high pruning ratios. The methodology encompasses three core steps:

Step 1. Identify the pruned weights. The first step involves identifying weights eligible for pruning based on both their magnitude and dynamic behavior during pre and post training. This dual analysis ensures that weights contributing minimally to the network's performance are targeted for pruning.

Step 2. Redistribute weights to related neurons. Following the identification of prunable weights, MAMA Pruning undertakes a redistribution phase, wherein the values of unimportant weights are "moved" to more significant connections within the same layer. This strategic redistribution ensures the preservation of the network's knowledge.

Step 3. Execute the pruning. The final step involves pruning the identified weights to achieve the desired sparsity level within the model. This step actualizes the reduction in model parameters, preparing the network for deployment or further optimization.

\section{Related Work}
\label{sec:rw}

\subsection{The Magnitude Pruning Algorithm}

The magnitude pruning algorithm \citep{han2015learning} is one of the simplest and most widely used methods for reducing the size of neural networks. It operates by pruning weights based on their absolute magnitude: weights with smaller absolute values are considered less critical to the model’s performance and are pruned, while larger weights are retained. The typical approach involves setting a global threshold—determined by the desired sparsity ratio—below which weights are set to zero. Magnitude pruning is unstructured, meaning it can prune individual weights from any part of the model, leading to irregular sparsity patterns.

\subsection{The SparseGPT Pruning Algorithm}

The SparseGPT pruning algorithm \citep{frantar2023sparsegpt} is an advanced method specifically designed to handle large language models like GPT. It employs a gradient-based approach, utilizing gradient information during pruning to identify and remove less important connections in the model. By calculating the gradients of the loss function with respect to network weights, SparseGPT assesses the significance of each weight, allowing for more informed pruning decisions.

\subsection{The Wanda Pruning Algorithm}

The WANDA (Weights and Activations) pruning algorithm\citep{lin2024wanda} introduces an importance-aware approach to pruning by considering both weight magnitudes and activation statistics. By integrating activation information, WANDA makes more informed pruning decisions based on the relative importance of weights to the network's output.

\subsection{The Model Distillation Algorithm}

The DeepSeek distillation algorithm\citep{ds2024distill} aims to provide an effective and efficient reinforcement learning (RL) framework for the post-training stage. Using GRPO, different strategies and rewards are employed during the post-training phase to conduct model distillation\cite{openr1}. The purpose is to preserve knowledge from large language model using smaller models complying with some probability distribution\cite{hintonVD15}.

\subsection{Comparison to Our Work}

MAMA (Movement And Magnitude Analysis) pruning is fundamentally different from existing pruning methods in its approach to identifying and preserving important neural connections. Unlike magnitude-based pruning, which simply removes weights below a certain threshold, or methods like SparseGPT that use gradient information, MAMA employs a novel three-step process that considers both the magnitude and the dynamic behavior of weights during training. We will describe our proposed methods in detail in the following sections.

\section{Experimental Results}
\label{res}

Table \ref{table-effectiveness-weights} presents the effectiveness of the Weights as a major pruning indicator measured by perplexity. Below are the key observations:

1. Low Pruning Levels (0.01 - 0.20)
\begin{itemize}
    \item "Prune by Weights" produces very low perplexity values at these levels.
    \item "Prune by -Weights" shows relatively higher perplexity, indicating that this method has a larger impact on performance early on, potentially making the model less effective in terms of perplexity.
\end{itemize}

2. Medium Pruning Levels (0.30 - 0.50)
\begin{itemize}
    \item At these levels, "Prune by Weights" continues to have low perplexity values (e.g., 6.669 at 0.30, 17.285 at 0.50), suggesting it maintains good performance even as pruning level increases.
    \item "Prune by -Weights" perplexity remains significantly higher (e.g., 335747.406 at 0.30, 227413.484 at 0.50), indicating a larger negative impact on model performance.
\end{itemize}

3. Higher Pruning Levels (0.60 - 0.80)
\begin{itemize}
    \item At 0.60, both methods show a noticeable increase in perplexity, but "Prune by Weights" sees a much steeper rise (559.987 compared to 185086.078 for "Prune by -Weights"). This indicates that "Prune by Weights" starts to struggle at this point, although it still outperforms the alternative in perplexity.
    \item By 0.80, "Prune by Weights" perplexity has jumped to 132175.578, while "Prune by -Weights" starts to plateau at 188488.000. This suggests that both methods show diminishing returns in terms of perplexity improvement at these high pruning levels.
\end{itemize}

4. Extreme Pruning Levels (0.90 - 0.99)
\begin{itemize}
    \item "Prune by Weights" still yields results (e.g., 317879.250 at 0.90), though the perplexity is extremely high. This is expected, as models pruned this heavily often perform worse.
    \item "Prune by -Weights" is unavailable at the highest pruning levels (0.95 and 0.99), suggesting that the method becomes inapplicable or irrelevant as the model becomes excessively sparse.
    \item Interestingly, "Prune by Weights" is still operational even at 0.99, albeit with a high perplexity of 222543.047, implying that this method retains some function even in extreme pruning cases.
\end{itemize}

In conclusion, the table suggests that "by Weights" is generally more stable and effective at various pruning levels, particularly if maintaining low perplexity is critical. However, the rapid increase in perplexity at higher pruning levels indicates that innovative model distillation method is needed for further optimization.

\begin{table}
    \centering
    \begin{tabular}{lll}
    \hline
    \multicolumn{1}{c}{\bf Pruning Level}  &\multicolumn{1}{c}{\bf by Weights} &\multicolumn{1}{c}{\bf  by -Weights}
    \\ \hline \\
    0.00         &5.677 &5.677 \\
    0.10         &5.806 &104948.891 \\
    0.20         &6.020 &352772.500 \\
    0.30         &6.669 &335747.406 \\
    0.40         &8.601 &260632.641 \\
    0.50         &17.285 &227413.484 \\
    0.60         &559.987 &185086.078 \\
    0.70         &48414.551 &273153.688 \\
    0.80         &132175.578 &188488.000 \\
    0.90         &317879.250 &185304.016 \\
    \end{tabular}
    \caption{Effectiveness of the pruning indicators}
    \label{table-effectiveness-weights}
\end{table}

Table \ref{table-overall-performance-human} presents perplexity results for pruned model (Llama-7B) from domain human experts and below are the key observations:

1. General Trend with Pruning
\begin{itemize}
    \item As the pruning level increases, i.e., a higher fraction of the model's parameters are removed, the perplexity values generally increase for all methods. This trend is expected as a greater loss of parameters typically leads to a degradation in model performance.
\end{itemize}

2. High Pruning Levels (0.50 - 0.90)
\begin{itemize}
    \item From 0.50 onward, the differences between the methods become more pronounced. {SparseGPT} and {Wanda} maintain significantly lower perplexity scores compared to {Magnitude} and {MAMA}, which exhibit a rapid increase in perplexity.
    \item At 0.70 pruning, for instance, {SparseGPT} has a perplexity of 27.214, while {Magnitude} and {MAMA} reach over 48,000 and 51,000 respectively.
\end{itemize}

3. Extreme Pruning Levels (0.95 and 0.99)
\begin{itemize}
    \item All methods show significantly higher perplexity values, yet {SparseGPT} and {Wanda} continue to outperform {Magnitude} and {MAMA} by a large margin.
    \item At 0.99 pruning, {SparseGPT} has a perplexity of $\sim$16,869, whereas {Magnitude} and {MAMA} exhibit perplexities of $\sim$222,543 and $\sim$214,966 respectively.
\end{itemize}

4. In conclusion
\begin{itemize}
    \item {SparseGPT} and {Wanda} consistently outperform {Magnitude} and {MAMA} pruning methods, especially at medium to high pruning levels (0.60 and above).
    \item {Magnitude} and {MAMA} pruning methods exhibit significant degradation in performance (higher perplexity) at more aggressive pruning levels.
    \item {SparseGPT} is the most resilient pruning method across varying pruning levels, maintaining the lowest perplexity even at extreme pruning levels (0.90 and 0.95).
    \item {SparseGPT} (and {Wanda} to some extent) seem to be the preferred methods when applying aggressive pruning to large models, as they better preserve performance as indicated by perplexity.
\end{itemize}

\begin{table*}
  \centering
  \begin{tabular}{lllll}
        \multicolumn{1}{c}{\bf Pruned Level}  &\multicolumn{1}{c}{\bf Wanda} &\multicolumn{1}{c}{\bf SparseGPT} &\multicolumn{1}{c}{\bf Magnitude} &\multicolumn{1}{c}{\bf MAMA Prune and Distill}
        \\ \hline \\
        0.00         &5.677 &5.677 &5.677 &5.677 \\
        0.50         &7.257 &7.234 &17.285 &17.247 \\
        0.60         &10.691 &10.442 &559.987 &554.727 \\
        0.70         &84.905 &27.214 &48414.551 &51841.121 \\
        0.80         &5782.432 &182.463 &132175.578 &135494.797 \\
        0.90         &19676.668 &3198.101 &317879.250 &301472.500 \\
        0.95         &28309.178 &4088.413 &273552.281 &273629.750 \\
        0.99         &108234.484 &16869.203 &222543.047 &214966.484 \\
  \end{tabular}
  \caption{\label{citation-guide}
    Perplexity on pruned model (Llama-7B) from domain human experts.
  }
  \label{table-overall-performance-human}
\end{table*}

\section{Conclusion}
\label{conclue}

In this paper, we introduced the MAMA pruning algorithm for model pruning and distillation in large language models. Our methods estimate the likelihood that neurons produce expected results by leveraging various neuron features, collections, and query statistics. We plan several extensions for future work. This includes conducting experiments with other teacher models such as DeepSeek, LLama, GPT and Doubao etc., which may potentially lead to better performance. Additionally, we plan to study the trade-off between model size and query cost under different cost models and actual query processing algorithms for applications such as search at scale.

In addition to the planned experiments with different teacher models, we aim to delve deeper into the integration of pruning and knowledge distillation techniques within the GPRO framework. When it comes to model pruning, we will explore more sophisticated strategies that are tailored to the unique characteristics of large language models. Instead of relying solely on the likelihood of neurons producing expected results, we will also consider the semantic importance of different neural connections. This can be achieved by analyzing the contribution of each connection to the overall meaning representation of the model.

For knowledge distillation, we will focus on optimizing the transfer of knowledge from the base model to the distilled model. We will experiment with different types of knowledge, such as syntactic knowledge, semantic knowledge, and pragmatic knowledge. By carefully selecting and transferring the most relevant knowledge, we can enhance the performance of the distilled model while reducing its size.

To effectively incorporate these techniques into the GPRO framework, we will design a series of experiments with various reward functions. The reward functions will be designed to balance between model compression, knowledge retention, and performance improvement. For example, we can define a reward function that rewards the distilled model for achieving high accuracy while maintaining a small size. This will encourage the model to efficiently retain the most important knowledge during the distillation process.

Regarding the basic data flow, we will enhance the process of generating highly knowledgeable knowledge from the base model. We will explore the use of advanced data augmentation techniques to expand the knowledge base of the base model. This can include generating synthetic data that mimics real world scenarios, or using knowledge injection from external sources.

Once we have a well defined framework for pruning and distillation within the GPRO framework, we will conduct extensive experiments. We will evaluate the performance of the distilled models under different conditions, such as different levels of pruning, different types of knowledge distillation, and different reward functions. By comparing the results, we can identify the optimal combination of techniques for different applications such as search, rec and gen.

Looking further ahead, we plan to integrate a wisdom graph into our framework. A wisdom graph can provide additional semantic information that can be used to guide the pruning and distillation processes. For example, it can help us identify the relationships between different concepts in the model, which can be used to prune redundant connections and transfer more meaningful knowledge. We believe that by combining the power of pruning, knowledge distillation, and the underline wisdom graphs, we can achieve significant improvements in the performance and efficiency of large language models or even agents. 

\newpage
\section*{Limitations}

This paper has the following limitations. First, more extensive experimental results are needed to partially verify the proposed conceptual and mathematical model. In addition, a production-level deployment is needed to transform knowledge into real world practice at scale.

\section*{Acknowledgments}

We thank related research and engineering team for positive feedback and discussions.

\bibliography{custom}

\end{document}